\documentclass[10pt,twocolumn,letterpaper]{article}

\usepackage{cvpr}              

\usepackage{graphicx}
\usepackage{amsmath}
\usepackage{amssymb}
\usepackage{booktabs}

\usepackage[pagebackref,breaklinks,colorlinks]{hyperref}

\usepackage[capitalize]{cleveref}
\crefname{section}{Sec.}{Secs.}
\Crefname{section}{Section}{Sections}
\Crefname{table}{Table}{Tables}
\crefname{table}{Tab.}{Tabs.}

\def\confName{CVPR}
\def\confYear{2022}

\begin{document}

\title{Analyzing VLM-Based Approaches for Anomaly Classification and Segmentation}


\author{
Mohit Kakda, Mirudula Shri Muthukumaran, Uttapreksha Patel, Lawrence Swaminathan Xavier Prince\\
Northeastern University\\
Boston, MA\\
{\tt\small \{kakda.m, muthukumaran.mi, patel.utt, swaminathanxavierp.l\}@northeastern.edu}
}
\maketitle

\begin{abstract}
Vision-Language Models (VLMs), particularly CLIP, have revolutionized anomaly detection by enabling zero-shot and few-shot defect identification without extensive labeled datasets. By learning aligned representations of images and text, VLMs facilitate anomaly classification and segmentation through natural language descriptions of normal and abnormal states, eliminating traditional requirements for task-specific training or defect examples. This project presents a comprehensive analysis of VLM-based approaches for anomaly classification (AC) and anomaly segmentation (AS). We systematically investigate key architectural paradigms including sliding window-based dense feature extraction (WinCLIP), multi-stage feature alignment with learnable projections (AprilLab framework), and compositional prompt ensemble strategies. Our analysis evaluates these methods across critical dimensions: feature extraction mechanisms, text-visual alignment strategies, prompt engineering techniques, zero-shot versus few-shot trade-offs, computational efficiency, and cross-domain generalization. Through rigorous experimentation on benchmarks such as MVTec AD and VisA, we compare classification accuracy, segmentation precision, and inference efficiency. The primary contribution is a foundational understanding of how and why VLMs succeed in anomaly detection, synthesizing practical insights for method selection and identifying current limitations. This work aims to facilitate informed adoption of VLM-based methods in industrial quality control and guide future research directions.
\end{abstract}

\vspace{-0.5em}\section{Introduction and Motivation}
\label{sec:intro}

Visual anomaly classification and segmentation are critical for automated industrial quality inspection, where systems must detect defects such as cracks, scratches, or misalignments in manufactured products. Traditional computer vision models rely on task-specific training, requiring large, annotated datasets for each defect type. This makes them costly, time-consuming, and difficult to scale when new defect types or product variations appear.

For instance, in a semiconductor manufacturing facility, conventional inspection systems must be retrained for every new circuit board design or defect type, such as missing components or burnt traces, delaying production and increasing labor costs. In contrast, vision-language models(VLMs)[17] can identify such anomalies by understanding natural language descriptions (e.g., “missing capacitor” or “discolored solder joint”), without requiring explicit retraining or defect examples.

Our project explores the use of pretrained vision-language models for zero-shot visual anomaly classification and segmentation, enabling defect detection using text prompts rather than labeled datasets. These models leverage aligned image–text understanding to generalize across unseen products and defect categories. We aim to analyze and evaluate how zero-shot and few-shot learning techniques can be applied to real-world inspection tasks, assessing their efficiency, adaptability, and accuracy. Ultimately, this work contributes toward developing scalable and generalizable inspection systems that minimize data requirements while improving the reliability and flexibility of industrial quality control.

Despite the growing number of VLM-based anomaly detection methods in recent literature[8,9,10], practitioners face uncertainty when selecting approaches for real-world deployment. Each method involves distinct trade-offs in accuracy, computational efficiency, interpretability, and ease of implementation. Our work addresses this gap by providing a systematic comparative analysis of prominent VLM-based techniques, identifying their strengths, limitations, and suitability for different inspection scenarios. This practical evaluation aims to guide engineers and researchers in making informed decisions when deploying these systems in production environments, considering constraints such as inference speed, hardware requirements, and defect type characteristics[6].

\subsection{Research Objectives}

This study focuses on two prominent VLM-based approaches that represent different design philosophies in anomaly detection:

\begin{itemize}
    \item \textbf{WinCLIP[8]}: Leverages sliding-window feature sampling and handcrafted compositional prompts, enabling zero-shot segmentation through explicit spatial reasoning.
    \item \textbf{AnomalyCLIP[14]}: Introduces learnable, object-agnostic prompts with multi-level optimization and DPAM, achieving better generalization and classification performance.
\end{itemize}

We aim to provide practitioners with a comprehensive understanding of their trade-offs in accuracy, computational cost, and deployment complexity.

\section{Related Work}

The field of anomaly detection has undergone significant evolution, with each generation of methods attempting to address the shortcomings of previous approaches while introducing new challenges of their own.

Early work in this domain focused heavily on reconstruction-based techniques. Methods using Autoencoders, Variational Autoencoders (VAEs)[16], and GAN[2]-based architectures operated on a straightforward premise: train a model to reconstruct normal patterns, and anomalies will naturally produce higher reconstruction errors. While intuitive, this approach has proven problematic in practice. These models tend to over-generalize, often reconstructing even defective regions with surprising accuracy, which undermines their ability to detect subtle anomalies. This limitation has motivated researchers to explore alternative strategies.

A different direction emerged with embedding-based methods like PatchCore[4], SPADE[5], and PaDiM[6]. Rather than reconstructing images, these approaches work within pretrained feature spaces to identify deviations from learned normal patterns. They excel at pixel-level localization and have shown impressive results on benchmark datasets. However, their reliance on extensive collections of normal samples makes them less flexible when dealing with new product types or unseen defect categories, a significant drawback in real manufacturing environments where product lines frequently change.

Self-supervised learning has offered another avenue forward. Techniques like CutPaste sidestep the need for real defect samples by creating synthetic anomalies through image augmentations. The model then learns to distinguish normal samples from these artificially perturbed versions using contrastive learning. While this removes the dependency on labeled anomalies, the approach hinges critically on the assumption that synthetic defects resemble real ones. When actual manufacturing defects look substantially different from the augmented examples, performance can suffer considerably.

More recently, Vision Language Models (VLMs) have opened up new possibilities for anomaly detection. Models like CLIP, which align visual and textual representations, enable zero-shot detection by matching images against natural language descriptions of defects. This paradigm shift is particularly appealing because it eliminates the need for retraining when encountering new defect types instead, one can simply describe the anomaly in text.

Our work focuses on comparing two prominent VLM-based approaches that exemplify different design philosophies. WinCLIP takes a relatively straightforward approach, using carefully designed prompt ensembles combined with window-based sampling of dense visual features. In contrast, AnomalyCLIP employs a more sophisticated pipeline involving object-agnostic prompt learning, multi-layer refinement of text tokens, and a specialized attention mechanism called Diagonally Prominent Attention Maps (DPAM) that enhances spatial localization. Both methods demonstrate strong generalization across diverse anomaly types and product categories, but they make different trade-offs in terms of complexity, computational cost, and detection accuracy. Understanding these trade-offs is essential for practitioners deciding which approach to deploy in real-world inspection systems.

\section{Method}

Our literature survey examined several prominent approaches for VLM-based anomaly detection, including WinCLIP, APRIL-GAN[9], CoOp[10], and AnomalyCLIP. After careful review, we identified WinCLIP and AnomalyCLIP as the most promising candidates for our comparative study. WinCLIP pioneered the application of vision-language models to industrial anomaly detection, introducing the concept of zero-shot defect classification through natural language prompts. AnomalyCLIP built upon this foundation by introducing learnable prompt engineering, eliminating the need for manually crafted object-specific descriptions. Both methods demonstrated strong theoretical advantages and empirical performance, but they take fundamentally different approaches to solving the same problem.

This section provides a detailed analysis of their architectural designs, highlighting the key differences that lead to their distinct performance characteristics. While both leverage CLIP's multimodal representation space, they differ substantially in how they construct textual prompts, extract visual features, and compute anomaly scores.

\subsection{WinCLIP}

WinCLIP was the first work to successfully adapt CLIP[16] for industrial anomaly detection. The core insight behind WinCLIP is that defects are often localized to small regions of an image, and a global image-level comparison might miss these subtle anomalies. To address this, the authors introduced a window-based approach that examines the image at a fine-grained level, combined with a carefully designed set of textual prompts that describe both normal and defective states.

\subsubsection{Sliding-Window Feature Extraction}

Rather than processing the entire image as a single entity, WinCLIP divides it into overlapping windows at multiple scales. For an input image $I$, this produces a set of windows:
\[
W = \{w_1, w_2, \ldots, w_N\}, \quad w_i \in \mathbb{R}^{h \times w}.
\]

Each window is passed independently through CLIP's image encoder to obtain dense patch-level embeddings:
\[
v_i = f_{\text{CLIP}}(w_i).
\]

The multi-scale nature of this approach is particularly clever. By using window sizes like $2 \times 2$, $3 \times 3$, and $5 \times 5$, WinCLIP can detect both small localized defects (captured by smaller windows) and larger structural anomalies (captured by larger windows). This dense extraction strategy gives the model a detailed, spatially-aware understanding of the image.

\subsubsection{Compositional Prompt Ensemble (CPE)}

One of WinCLIP's main contributions is its systematic approach to prompt design. Instead of using a single generic prompt, it constructs an ensemble of carefully crafted descriptions organized into two categories:

\begin{itemize}
    \item \textbf{Normality prompts:} ``flawless [object],'' ``perfect [object],'' ``unblemished [object].''
    \item \textbf{Anomaly prompts:} ``damaged [object],'' ``[object] with defect,'' ``broken [object].''
\end{itemize}

These base prompts are then expanded using various template phrases to create richer descriptions:
\begin{itemize}
    \item ``a photo of a [state] for visual inspection''
    \item ``a close-up photo of the [state]''
    \item ``a cropped photo of a [state]''
\end{itemize}

The text encoder processes each prompt $t_j$ to obtain textual embeddings:
\[
u_j = g_{\text{CLIP}}(t_j).
\]

Classification happens by measuring similarity between each window embedding and all prompt embeddings. For each window, the model computes:
\[
s_i = \max_j \cos(v_i, u_j).
\]

The final pixel-level anomaly map is generated by interpolating these window-level scores back to the original image resolution. This gives a spatial heatmap showing which regions are most likely to contain defects.

\subsubsection{What Works Well and What Doesn't}

WinCLIP's window-based approach has clear advantages. Because each window contributes independently to the anomaly map, the method excels at localizing small defects. The compositional prompt ensemble also provides robustness by capturing different ways of describing the same concept.

However, the method has some notable limitations. First, designing effective prompts requires domain knowledge about the specific objects being inspected. A prompt that works well for detecting scratches on metal surfaces might not transfer well to defects in textiles. Second, the sliding-window evaluation is computationally expensive, especially when using multiple scales. Finally, because each window is processed in isolation, WinCLIP can struggle with anomalies that require understanding global structure or context, for example, detecting that a component is missing from an assembly, or that parts are misaligned relative to each other.

\subsection{AnomalyCLIP}

AnomalyCLIP[14] takes a fundamentally different philosophical approach. Rather than manually engineering prompts for each object category, it learns generic textual representations that capture the universal concept of ``abnormality.'' This shift from handcrafted to learned prompts is significant because it makes the method truly zero-shot, it can detect defects in entirely new object categories without any prompt engineering effort.

\subsubsection{Object-Agnostic Prompt Learning}

The key innovation here is replacing fixed text prompts with learnable tokens. AnomalyCLIP initializes a set of trainable embeddings:
\[
T = \{t_1, t_2, \ldots, t_k\}.
\]

These tokens are inserted into multiple layers of CLIP's text encoder, and each layer learns to refine them:
\[
h_\ell = \text{TransformerLayer}_\ell(h_{\ell-1}, T_\ell),
\]
where $T_\ell$ represents the layer-specific learned tokens.

This multi-level refinement is important because different transformer layers capture different types of information. Early layers might learn low-level cues like texture irregularities, while deeper layers capture high-level semantic concepts like structural inconsistencies. By learning prompts at multiple levels, the model can encode a rich, hierarchical understanding of what makes something anomalous.

\subsubsection{Unified Global and Local Optimization}

Unlike WinCLIP, which focuses primarily on producing good anomaly maps, AnomalyCLIP optimizes for both image-level classification and pixel-level segmentation simultaneously. This is formalized through a combined loss function.

For classification, a standard cross-entropy loss is used:
\[
L_{\text{global}} = \text{CE}(y, \hat{y}).
\]

For segmentation, the model uses a combination of focal loss and dice loss to handle class imbalance and ensure precise boundary delineation:
\[
L_{\text{local}} = \lambda_1 \cdot \text{FocalLoss} + \lambda_2 \cdot \text{DiceLoss}.
\]

The total training objective becomes:
\[
L = L_{\text{global}} + \sum_{p \in \text{pixels}} L_{\text{local}}(p).
\]

This joint optimization strategy has a nice practical benefit: the model learns to both correctly classify whether an image contains any defects (global task) and precisely locate where those defects are (local task). This makes AnomalyCLIP particularly effective for subtle anomalies that might be easy to overlook if you're only looking at coarse image-level features.

\subsubsection{Diagonally Prominent Attention Maps (DPAM)}

One of the more technical but important contributions of AnomalyCLIP is its modification to CLIP's attention mechanism. In standard CLIP, attention weights tend to focus on semantically rich or visually salient regions. While this works well for general image understanding, it can cause the model to overlook small or low-contrast defects.

AnomalyCLIP addresses this by amplifying each token's self-attention:
\[
A' = A + \gamma \cdot \text{diag}(A),
\]
where $A$ is the original attention matrix and $\gamma$ is a learned scaling factor.

This modification encourages the model to pay more attention to local features, which turns out to be crucial for detecting:
\begin{itemize}
    \item tiny defects that occupy only a few pixels,
    \item low-contrast anomalies that are barely visible,
    \item subtle texture variations that differ only slightly from normal patterns.
\end{itemize}

\subsubsection{Why AnomalyCLIP Works}

AnomalyCLIP's strengths come from several factors working together. The learned prompts provide true zero-shot generalization across object categories. The unified training objective ensures both accurate classification and precise segmentation. The DPAM mechanism specifically addresses the challenge of detecting small, subtle anomalies. Combined, these design choices make AnomalyCLIP particularly well-suited for real-world industrial inspection, where defects vary widely in size, contrast, and appearance.

\section{Experiments and Results}
\label{sec:experiments}

\subsection{Dataset and Experimental Setup}

Since the aim of this paper is to evaluate methods for industrial-grade anomaly detection, we use the MVTec AD dataset[12] to compare WinCLIP and AnomalyCLIP. The MVTec AD dataset provides a comprehensive testbed with 15 diverse object categories ranging from rigid objects (bottle, capsule, metal nut, screw, transistor) to textures (carpet, grid, leather, tile, wood) and flexible objects (cable, zipper). Each category contains multiple defect types; for example, the bottle category includes subcategories such as 'contamination', 'broken large', and 'broken small'. This diversity in object types and defect patterns makes MVTec AD an ideal benchmark for evaluating zero-shot anomaly detection methods.

\subsection{Implementation Details}

\textbf{WinCLIP:} We evaluate WinCLIP across multiple scenarios: zero-shot, one-shot, and few-shot settings. For the zero-shot scenario, the model uses only template-based prompts without any training samples. In the one-shot and few-shot scenarios, the model is provided with 1 and 4 normal reference images respectively. Due to space constraints, we present the zero-shot results in the main paper and defer the one-shot and few-shot results to the appendix.

\textbf{AnomalyCLIP:} We use the official implementation from \url{https://github.com/zqhang/AnomalyCLIP}. Following the original paper, the visual encoder weights from CLIP are kept frozen, while learnable tokens are introduced at each layer of the text encoder. Importantly, the learnable tokens from previous layers are discarded at each subsequent layer, implementing the multi-layer text refinement strategy. We use the pre-trained weights provided by the authors to ensure reproducibility and fair comparison.

\subsection{Evaluation Metrics}

We evaluate both methods using four standard metrics:
\begin{itemize}
    \item \textbf{Classification AUROC}: Area under the ROC curve for image-level anomaly classification
    \item \textbf{Classification AP}: Average precision for image-level classification
    \item \textbf{Segmentation AUROC}: Pixel-level anomaly localization performance
    \item \textbf{Segmentation AUPRO}: Area under the per-region overlap curve, emphasizing localization quality
\end{itemize}

\subsection{Results}

Tables~\ref{tab:winclip_results} and~\ref{tab:anomalyclip_results} present the detailed per-category performance of WinCLIP and AnomalyCLIP respectively on the MVTec AD dataset. Tables~\ref{tab:winclip_summary} and~\ref{tab:anomalyclip_summary} summarize the average performance across all categories.

\begin{table}[t]
\centering
\caption{WinCLIP Performance on MVTec AD Dataset}
\label{tab:winclip_results}
\begin{tabular}{lcc}
\toprule
\textbf{Category} & \textbf{Classification AUROC} & \textbf{Segmentation AUROC} \\
\midrule
Bottle & 0.892 & 0.553 \\
Cable & 0.507 & 0.585 \\
Capsule & 0.506 & 0.811 \\
Carpet & 0.847 & 0.915 \\
Grid & 0.591 & 0.762 \\
Hazelnut & 0.904 & 0.799 \\
Leather & 0.563 & 0.848 \\
Metal Nut & 0.732 & 0.624 \\
Pill & 0.569 & 0.749 \\
Screw & 0.537 & 0.853 \\
Tile & 0.149 & 0.739 \\
Toothbrush & 0.567 & 0.731 \\
Transistor & 0.575 & 0.547 \\
Wood & 0.677 & 0.751 \\
Zipper & 0.575 & 0.632 \\
\midrule
\textbf{Average} & \textbf{0.612} & \textbf{0.726} \\
\bottomrule
\end{tabular}
\end{table}

\begin{table}[t]
\centering
\caption{WinCLIP Average Performance Across All MVTec AD Categories}
\label{tab:winclip_summary}
\begin{tabular}{lc}
\toprule
\textbf{Metric} & \textbf{Average Score} \\
\midrule
Classification AUROC & 0.612 \\
Classification AP & 0.806 \\
Segmentation AUROC & 0.726 \\
Segmentation AUPRO & 0.679 \\
\bottomrule
\end{tabular}
\end{table}

\begin{table}[t]
\centering
\caption{AnomalyCLIP Performance on MVTec AD Dataset}
\label{tab:anomalyclip_results}
\begin{tabular}{lcc}
\toprule
\textbf{Category} & \textbf{Classification AUROC} & \textbf{SegmentationAUROC} \\
\midrule
Bottle & 0.887 & 0.904 \\
Cable & 0.703 & 0.789 \\
Capsule & 0.895 & 0.958 \\
Carpet & \textbf{1.000} & 0.988 \\
Grid & 0.978 & 0.973 \\
Hazelnut & 0.972 & 0.972 \\
Leather & 0.998 & 0.986 \\
Metal Nut & 0.924 & 0.746 \\
Pill & 0.811 & 0.918 \\
Screw & 0.821 & 0.975 \\
Tile & \textbf{1.000} & 0.947 \\
Toothbrush & 0.853 & 0.919 \\
Transistor & 0.939 & 0.708 \\
Wood & 0.969 & 0.964 \\
Zipper & 0.984 & 0.913 \\
\midrule
\textbf{Average} & \textbf{0.916} & \textbf{0.907} \\
\bottomrule
\end{tabular}
\end{table}

\begin{table}[t]
\centering
\caption{AnomalyCLIP Average Performance Across All MVTec AD Categories}
\label{tab:anomalyclip_summary}
\begin{tabular}{lc}
\toprule
\textbf{Metric} & \textbf{Average Score} \\
\midrule
Classification AUROC & 0.916 \\
Classification AP & 0.963 \\
Segmentation AUROC & 0.907 \\
Segmentation AUPRO & 0.814 \\
\midrule
Total Test Images & 1,725 \\
\bottomrule
\end{tabular}
\end{table}

\subsection{Performance Analysis}

\textbf{Overall Performance and Consistency:} AnomalyCLIP demonstrates substantially superior and more consistent performance compared to WinCLIP across all evaluation metrics. AnomalyCLIP achieves classification AUROC of 0.916 versus WinCLIP's 0.612 (49.7\% improvement) and segmentation AUROC of 0.907 versus 0.726 (24.9\% improvement). Beyond higher performance, AnomalyCLIP exhibits lower variance across categories, classification AUROC ranges from 0.703 (cable) to 1.000 (carpet, tile) compared to WinCLIP's wider range of 0.149 (tile) to 0.904 (hazelnut). This consistency is crucial for practical deployment where reliability across diverse scenarios is essential.

\begin{figure}[t]
\centering
\includegraphics[width=\columnwidth]{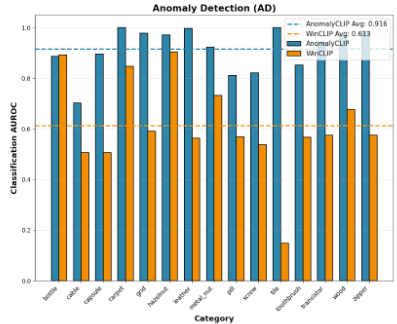}
\caption{Classification performance (AUROC) comparison between AnomalyCLIP and WinCLIP across all 15 MVTec AD categories. Dashed lines indicate average performance (AnomalyCLIP: 0.916, WinCLIP: 0.612).}
\label{fig:classification}
\end{figure}

\textbf{Texture vs. Object Categories:} AnomalyCLIP shows particularly strong performance on texture-based categories. For carpet and tile, it achieves perfect classification AUROC scores of 1.000, with segmentation AUROC of 0.988 and 0.947 respectively. Similarly, leather (0.998 classification AUROC, 0.986 segmentation AUROC) and wood (0.969 classification AUROC, 0.964 segmentation AUROC) demonstrate excellent performance. This suggests that object-agnostic learning is especially effective for texture anomalies where the distinction between normal and abnormal patterns does not depend on specific object semantics.

\textbf{WinCLIP's Window-Based Advantages:} Despite being numerically inferior in overall performance, WinCLIP exhibits interesting strengths in specific scenarios. Notably, WinCLIP achieves competitive segmentation performance on categories with fine structural details, such as transistors (segmentation AUPRO of 0.704) and screws (segmentation AUROC of 0.853). The window-based approach appears particularly effective at identifying missing components and detecting small-scale anomalies. This suggests that for industries such as semiconductor manufacturing, circuit board inspection, and mechanical component quality control, where precise localization of minute defects is critical, a window-based approach may yield more practical results despite lower overall classification scores.

\textbf{Challenging Categories:} Both methods struggle with certain categories, albeit to different degrees. Cable presents challenges for both approaches, with AnomalyCLIP achieving 0.703 classification AUROC and WinCLIP only 0.507. Interestingly, for some categories like transistor and metal nut, AnomalyCLIP shows relatively weaker segmentation performance (0.708 and 0.746 respectively) despite strong classification scores. This may indicate difficulties in precise anomaly localization for complex structural objects with intricate geometries.

\begin{figure}[t]
\centering
\includegraphics[width=\columnwidth]{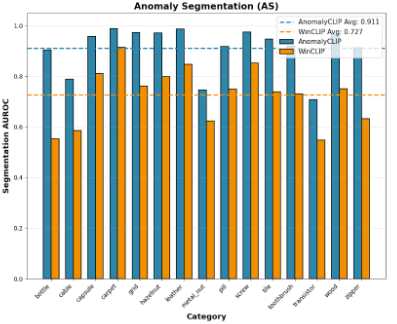}
\caption{Segmentation performance (AUROC) comparison between AnomalyCLIP and WinCLIP across all 15 MVTec AD categories. Dashed lines indicate average performance (AnomalyCLIP: 0.907, WinCLIP: 0.726).}
\label{fig:segmentation}
\end{figure}

\textbf{Segmentation Performance Gap:} While AnomalyCLIP outperforms WinCLIP in classification by a large margin (0.916 vs 0.612), the segmentation gap is smaller (0.907 vs 0.726). This suggests that WinCLIP's window-based approach, despite weaker classification, provides reasonable spatial localization capabilities. The explicit spatial windowing mechanism enables fine-grained anomaly localization that partially compensates for its weaker semantic understanding. However, AnomalyCLIP still maintains a significant advantage in pixel-level anomaly localization while simultaneously providing superior classification performance.

\textbf{Localization Quality (AUPRO):} The AUPRO metric, which emphasizes connected component overlap, reveals important insights. AnomalyCLIP achieves an average AUPRO of 0.814, while WinCLIP achieves 0.679. The smaller gap compared to AUROC (0.907 vs 0.726) suggests that both methods achieve reasonable region-level localization, though pixel-level precision differs. Notably, categories like hazelnut (0.925), leather (0.922), and wood (0.915) show excellent AUPRO scores for AnomalyCLIP, indicating precise defect boundary delineation.

\textbf{Performance by Object Type:} We observe distinct performance patterns across different object types. 

\textit{Texture objects} (carpet, grid, leather, tile, wood): AnomalyCLIP achieves exceptional performance with an average classification AUROC of 0.989 and segmentation AUROC of 0.972. WinCLIP performs moderately with averages of 0.565 and 0.803 respectively.

\textit{Rigid objects} (bottle, capsule, hazelnut, metal nut, pill, screw, toothbrush): AnomalyCLIP maintains strong performance (avg. 0.870 classification, 0.901 segmentation), while WinCLIP shows more variability (avg. 0.684 classification, 0.711 segmentation).

\textit{Flexible/Complex objects} (cable, transistor, zipper): Both methods struggle, with AnomalyCLIP achieving 0.875 average classification and 0.803 segmentation, and WinCLIP achieving 0.552 and 0.588 respectively.

\textbf{Failure Case Analysis:} 
\textit{Cable (0.703/0.507 classification):} The flexible, varying shape and thin structure make it difficult to establish a consistent "normal" appearance. Defects often involve subtle bending or kinking that both methods struggle to detect.

\textit{Tile (WinCLIP 0.149):} WinCLIP's dramatic failure on tile likely stems from its template-based prompts being poorly suited for subtle texture variations. The uniform appearance makes window-based discrimination challenging.

\textit{Transistor segmentation (AnomalyCLIP 0.708):} Despite strong classification, precise localization is difficult due to dense, repetitive component structures where small missing parts are hard to segment accurately.

\section{Discussion}
\label{sec:discussion}





\subsection{AnomalyCLIP Deep Dive: Performance Patterns}

\subsubsection{Category-Type Analysis}

We analyze AnomalyCLIP's performance across three distinct object types to understand where object-agnostic learning excels:

\textbf{Texture Objects} (carpet, grid, leather, tile, wood): AnomalyCLIP achieves exceptional performance with an average classification AUROC of 0.989 and segmentation AUROC of 0.972. The highly regular patterns and texture-based anomalies in these categories align well with object-agnostic prompt learning, where semantic object information is less critical than appearance consistency.

\textbf{Rigid Objects} (bottle, capsule, hazelnut, metal nut, pill, screw, toothbrush): AnomalyCLIP maintains strong performance with average classification AUROC of 0.870 and segmentation AUROC of 0.901. The structural nature of anomalies in these categories benefits from CLIP's visual understanding combined with learned anomaly-specific representations.

\textbf{Flexible/Complex Objects} (cable, transistor, zipper): Both methods struggle with these categories, where AnomalyCLIP achieves 0.875 average classification AUROC and 0.803 segmentation AUROC. The variable shapes and complex structures make it challenging to establish consistent normal patterns, highlighting a limitation of current zero-shot approaches.

\subsubsection{Defect Characteristic Analysis}

Beyond category-level analysis, we investigate how different defect characteristics impact AnomalyCLIP's detection performance using MVTec AD's rich defect annotations.

\begin{table}[t]
\centering
\caption{AnomalyCLIP Performance by Defect Visual Contrast}
\label{tab:defect_contrast}
\small
\begin{tabular}{lccc}
\toprule
\textbf{Contrast Level} & \textbf{AUROC} & \textbf{Std Dev} & \textbf{Count} \\
\midrule
Low & 0.768 & 0.090 & 13 \\
Medium & 0.900 & 0.098 & 14 \\
High & 0.939 & 0.101 & 25 \\
\bottomrule
\end{tabular}
\end{table}

\textbf{Visual Contrast Analysis:} Table~\ref{tab:defect_contrast} reveals a clear correlation between defect visual contrast and detection performance. High-contrast defects (AUROC 0.939) are detected significantly more reliably than low-contrast defects (AUROC 0.768), representing a 17.1 percentage point gap. This highlights a fundamental limitation of vision-based anomaly detection: subtle defects that are visually similar to normal regions remain challenging. The medium-contrast category (AUROC 0.900) suggests a threshold effect where sufficient visual distinction enables reliable detection. This has practical implications for quality control applications, AnomalyCLIP is most suitable for scenarios where defects manifest as clear visual deviations rather than subtle appearance changes.

\begin{table}[t]
\centering
\caption{AnomalyCLIP Performance by Defect Type Category}
\label{tab:defect_type}
\small
\begin{tabular}{lccc}
\toprule
\textbf{Defect Type} & \textbf{AUROC} & \textbf{Std} & \textbf{N} \\
\midrule
Missing Component & 0.757 & 0.173 & 3 \\
Logical Error & 0.801 & 0.092 & 3 \\
Geometric Deformation & 0.808 & 0.197 & 3 \\
Surface Pattern & 0.813 & 0.093 & 2 \\
Multiple & 0.866 & 0.090 & 4 \\
Surface Damage & 0.880 & 0.111 & 20 \\
Foreign Material & 0.881 & 0.144 & 4 \\
Appearance Change & 0.967 & 0.051 & 5 \\
Structural Break & 0.983 & 0.022 & 5 \\
Missing Material & 0.999 & 0.001 & 3 \\
\bottomrule
\end{tabular}
\end{table}

\textbf{Defect Type Characteristics:} Table~\ref{tab:defect_type} provides critical insights into which defect types are most challenging. \textit{Missing components} (AUROC 0.757) and \textit{logical errors} (AUROC 0.801) represent the most difficult categories, likely because they require understanding of functional correctness rather than purely visual appearance. In contrast, \textit{missing material} (AUROC 0.999), \textit{structural breaks} (AUROC 0.983), and \textit{appearance changes} (AUROC 0.967) are detected with high reliability as they produce clear visual signatures. \textit{Surface damage} (AUROC 0.880), being the most common defect type (n=20), represents the typical performance users can expect in industrial scenarios. The high standard deviation for geometric deformation (0.197) suggests that some deformations are obvious while others are subtle, depending on severity and viewing angle.

\textbf{Key Insights:} Our defect-level analysis reveals that AnomalyCLIP excels at detecting high-contrast defects with clear visual signatures (structural breaks, missing material, appearance changes) but struggles with low-contrast defects and those requiring functional understanding (missing components, logical errors). This characterization provides practitioners with clear guidance on deployment suitability: AnomalyCLIP is ideal for visual quality inspection but may require complementary methods for functional testing or detecting subtle defects in low-contrast scenarios.

\subsection{WinCLIP Ablation Studies}

We conduct comprehensive ablation studies to understand WinCLIP's behavior under different configurations and how few-shot learning impacts its performance.

\subsubsection{Few-Shot Learning Impact}

\begin{figure}[t]
\centering
\includegraphics[width=\columnwidth]{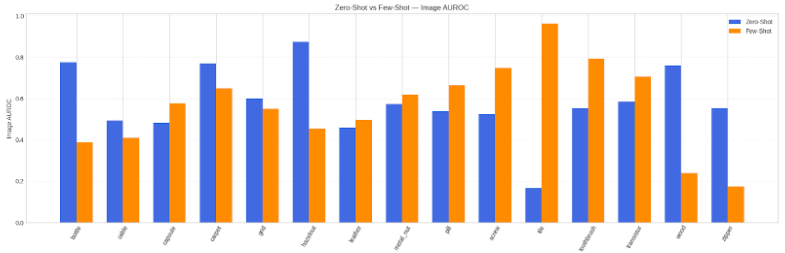}
\caption{Direct comparison of WinCLIP zero-shot vs few-shot (4-shot) performance on image-level anomaly detection (AUROC) across MVTec AD categories.}
\label{fig:zero_vs_few_image}
\end{figure}

Figure 3 and Figure 4 presents WinCLIP's performance across different shot scenarios. Several important patterns emerge:

\textbf{Few-shot learning patterns:} Most categories show performance gains when moving from zero-shot to few-shot settings, with the largest improvements occurring between zero-shot and 1-shot (average +0.15 AUROC). Categories like capsule, grid, and hazelnut demonstrate substantial improvements of 0.2-0.3 AUROC with just one reference image, while carpet and hazelnut achieve near-perfect performance (\(>0.95\) AUROC) in few-shot settings. However, diminishing returns appear beyond 1-2 samples, suggesting that a single well-chosen reference image captures most of the normal appearance characteristics for window-based comparison.






\begin{figure}[t]
\centering
\includegraphics[width=\columnwidth]{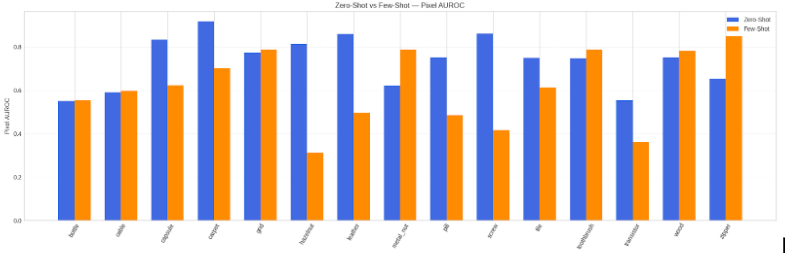}
\caption{Direct comparison of WinCLIP zero-shot vs few-shot (4-shot) performance on pixel-level anomaly detection (AUROC) across MVTec AD categories.}
\label{fig:zero_vs_few_image}
\end{figure}


\textbf{Image-level detection:} Few-shot learning provides substantial gains for most categories, with tile showing the most dramatic improvement (from 0.15 to 0.95 AUROC). However, some categories like bottle and screw show surprising performance drops, suggesting that the selected reference images may introduce noise or that template-based prompts were already optimal.

\textbf{Pixel-level segmentation:} The pattern is more nuanced at the pixel level. While categories like capsule, grid, and wood show clear improvements, others like hazelnut and leather experience slight degradation. This indicates that reference samples help with overall anomaly detection but may not always improve precise localization, possibly due to window-level misalignments or variations in defect appearance.

\subsubsection{Window Size Analysis}

\begin{figure}[t]
\centering
\includegraphics[width=\columnwidth]{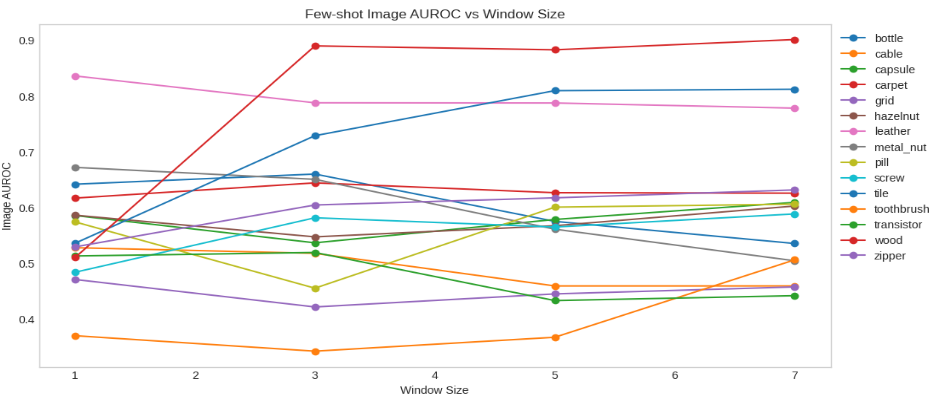}
\caption{Impact of window size on WinCLIP's few-shot image-level anomaly detection performance (AUROC) across all MVTec AD categories. Window sizes range from 1 (single pixel) to 7.}
\label{fig:window_size}
\end{figure}

Figure~\ref{fig:window_size} reveals the critical impact of window size selection on WinCLIP's performance. Different categories exhibit distinct optimal window sizes based on their defect characteristics: texture objects (carpet, grid, leather) perform best with intermediate sizes (3-5) as they need sufficient context to assess pattern regularity without diluting local texture variations, while structural objects (bottle, capsule) show monotonic improvement with larger windows (5-7) because their defects (cracks, deformations) often span broader spatial regions. Carpet exemplifies texture sensitivity, with dramatic performance variation from 0.62 (size 1) to 0.89 (size 3) before declining at larger sizes. Categories with fine-grained components like transistor would theoretically benefit from smaller windows to detect missing parts, though our results show relative stability across sizes, possibly due to the challenge of detecting subtle component-level defects regardless of window size. Window size 1 performs universally poorly (average AUROC $\sim$0.55), confirming that spatial context is essential for reliable anomaly detection.

\textbf{Practical implications:} For deployment, we recommend starting with window size 3-5 as a reasonable default, using 5-7 for large structural objects where defects span regions (wood grain defects, bottle cracks), using 3-4 for fine-detail inspection (PCB components, small mechanical parts), and employing cross-validation to tune per application. The category-dependent optimal window size reveals a fundamental trade-off: smaller windows provide attention to fine details but lack contextual information, while larger windows capture broader patterns but may average out localized defects.

\subsection{Limitations and Future Work}

\textbf{Current Limitations:} Both methods struggle with flexible objects (cable), low-contrast defects, and have not been extensively validated in real-world industrial conditions with varying lighting and viewing angles. AnomalyCLIP shows weakness in missing component detection and logical errors, while WinCLIP exhibits high cross-category variance.

\textbf{Hybrid Architecture:} A promising direction is combining AnomalyCLIP's object-agnostic learnable prompts with WinCLIP's sliding window architecture. This hybrid approach could leverage semantic anomaly understanding from learned prompts while maintaining explicit spatial reasoning for fine-grained localization. Specifically, AnomalyCLIP's text encoder could generate anomaly-aware representations, which are then applied within a sliding window framework for pixel-level scoring. This addresses AnomalyCLIP's localization weaknesses (e.g., transistor segmentation AUROC 0.708) while maintaining robust classification performance.

\textbf{Learning from Misclassifications:} Systematic analysis of false positives and false negatives could reveal critical failure patterns. By characterizing when and why models fail, whether due to lighting variations, ambiguous boundaries, or specific defect types we can develop targeted improvements through hard negative mining, contrastive learning on misclassified samples, or specialized modules for problematic categories. This iterative refinement could significantly improve deployment reliability, particularly for challenging low-contrast and missing-component scenarios.

\textbf{Other Directions:} Multi-scale fusion across window sizes, temporal consistency for video inspection, domain adaptation to specialized applications (medical, semiconductor), computational optimization for real-time deployment, and explainability tools for human-in-the-loop quality control.

\section{Conclusion}

This work compared WinCLIP and AnomalyCLIP to help practitioners choose the right approach for industrial anomaly detection. Our evaluation on MVTec AD revealed that AnomalyCLIP substantially outperforms WinCLIP, achieving 91.6\% classification AUROC versus 61.2\%, and 90.7\% segmentation AUROC versus 72.6\%. AnomalyCLIP's learnable prompts and DPAM mechanism provide superior generalization across object types without manual engineering. However, WinCLIP offers advantages in truly zero-shot deployment and explicit spatial reasoning for fine-grained localization.

The choice depends on deployment constraints: AnomalyCLIP excels where accuracy matters most, while WinCLIP suits rapid prototyping with limited resources. Both methods struggle with flexible objects and complex assemblies, indicating promising directions for future research. Overall, vision-language models have fundamentally changed industrial inspection, enabling defect detection through natural language without extensive retraining.







{
    \small
}

[1] Rumelhart, D.E., Hinton, G.E. and Williams, R.J., 1986. Learning representations by back-propagating errors. nature, 323(6088), pp.533-536.

[2] Goodfellow, I.J., Pouget-Abadie, J., Mirza, M., Xu, B., Warde-Farley, D., Ozair, S., Courville, A. and Bengio, Y., 2014. Generative adversarial nets. Advances in neural information processing systems, 27.

[3] Zavrtanik, V., Kristan, M. and Skočaj, D., 2021. Draem-a discriminatively trained reconstruction embedding for surface anomaly detection. In Proceedings of the IEEE/CVF international conference on computer vision (pp. 8330-8339).

[4] Roth, K., Pemula, L., Zepeda, J., Schölkopf, B., Brox, T. and Gehler, P., 2022. Towards total recall in industrial anomaly detection. In Proceedings of the IEEE/CVF conference on computer vision and pattern recognition (pp. 14318-14328).

[5] Park, T., Liu, M.Y., Wang, T.C. and Zhu, J.Y., 2019. Semantic image synthesis with spatially-adaptive normalization. In Proceedings of the IEEE/CVF conference on computer vision and pattern recognition (pp. 2337-2346).

[6] Defard, T., Setkov, A., Loesch, A. and Audigier, R., 2021, January. Padim: a patch distribution modeling framework for anomaly detection and localization. In International conference on pattern recognition (pp. 475-489). Cham: Springer International Publishing.

[7] Li, C.L., Sohn, K., Yoon, J. and Pfister, T., 2021. Cutpaste: Self-supervised learning for anomaly detection and localization. In Proceedings of the IEEE/CVF conference on computer vision and pattern recognition (pp. 9664-9674).

[8] Jeong, J., Zou, Y., Kim, T., Zhang, D., Ravichandran, A. and Dabeer, O., 2023. Winclip: Zero-/few-shot anomaly classification and segmentation. In Proceedings of the IEEE/CVF Conference on Computer Vision and Pattern Recognition (pp. 19606-19616).

[9] Chen, X., Han, Y. and Zhang, J., 2023. April-gan: A zero-/few-shot anomaly classification and segmentation method for cvpr 2023 vand workshop challenge tracks 1\&2: 1st place on zero-shot ad and 4th place on few-shot ad. arXiv preprint arXiv:2305.17382.

[10] Zhou, K., Yang, J., Loy, C.C. and Liu, Z., 2022. Learning to prompt for vision-language models. International Journal of Computer Vision, 130(9), pp.2337-2348.

[11] Huang, C., Guan, H., Jiang, A., Zhang, Y., Spratling, M. and Wang, Y.F., 2022, October. Registration based few-shot anomaly detection. In European conference on computer vision (pp. 303-319). Cham: Springer Nature Switzerland.

[12] Bergmann, P., Fauser, M., Sattlegger, D. and Steger, C., 2019. MVTec AD--A comprehensive real-world dataset for unsupervised anomaly detection. In Proceedings of the IEEE/CVF conference on computer vision and pattern recognition (pp. 9592-9600).

[13] Zou, Y., Jeong, J., Pemula, L., Zhang, D. and Dabeer, O., 2022, October. Spot-the-difference self-supervised pre-training for anomaly detection and segmentation. In European conference on computer vision (pp. 392-408). Cham: Springer Nature Switzerland.

[14] Zhou, Q., Pang, G., Tian, Y., He, S. and Chen, J., 2023. Anomalyclip: Object-agnostic prompt learning for zero-shot anomaly detection. arXiv preprint arXiv:2310.18961.

[15] Kingma, D.P. and Welling, M., 2013. Auto-encoding variational bayes. arXiv preprint arXiv:1312.6114.

[16] Radford, A., Kim, J.W., Hallacy, C., Ramesh, A., Goh, G., Agarwal, S., Sastry, G., Askell, A., Mishkin, P., Clark, J. and Krueger, G., 2021, July. Learning transferable visual models from natural language supervision. In International conference on machine learning (pp. 8748-8763). PmLR.

[17] Bordes, F., Pang, R.Y., Ajay, A., Li, A.C., Bardes, A., Petryk, S., Mañas, O., Lin, Z., Mahmoud, A., Jayaraman, B. and Ibrahim, M., 2024. An introduction to vision-language modeling. arXiv preprint arXiv:2405.17247.

\end{document}